\documentclass[10pt,twocolumn,letterpaper]{article}

\usepackage{cvpr}              % To produce the CAMERA-READY version
\usepackage{graphicx}
\usepackage{amsmath}
\usepackage{amssymb}
\usepackage{booktabs}

\usepackage{epsfig}
\usepackage{graphicx}
\usepackage{amsmath}
\usepackage{amssymb}
\usepackage{multirow}
\usepackage{enumerate}
\usepackage{bm}
\usepackage{wrapfig}
\usepackage{epstopdf}
\usepackage{wrapfig}

\usepackage{cuted}
\usepackage{capt-of}

\usepackage[pagebackref,breaklinks,colorlinks]{hyperref}

\usepackage[capitalize]{cleveref}
\crefname{section}{Sec.}{Secs.}
\Crefname{section}{Section}{Sections}
\Crefname{table}{Table}{Tables}
\crefname{table}{Tab.}{Tabs.}

\def\cvprPaperID{4834} % *** Enter the CVPR Paper ID here
\def\confName{CVPR}
\def\confYear{2023}

\begin{document}

\title{Towards Better Gradient Consistency for Neural Signed Distance Functions via Level Set Alignment}

\author{Baorui Ma$^{1}$$\thanks{Equal contribution.}$, Junsheng Zhou$^{1*}$, Yu-Shen Liu$^{1}$\thanks{The corresponding author is Yu-Shen Liu. This work was supported by National Key R\&D Program of China (2022YFC3800600), the National Natural Science Foundation of China (62272263, 62072268), and in part by Tsinghua-Kuaishou Institute of Future Media Data.}, Zhizhong Han$^2$\\
%\affiliations
$^1$School of Software, BNRist, Tsinghua University, Beijing, China\\
$^2$Department of Computer Science, Wayne State University, Detroit, USA\\
%\emails
{\tt\small \{mbr18,zhoujs21\}@mails.tsinghua.edu.cn,  liuyushen@tsinghua.edu.cn, h312h@wayne.edu}
}

\maketitle

%\begin{strip}\centering
%\vspace{-0.70in}
%\includegraphics[width=\textwidth]{../Figures/Paris1.eps}
%\vspace{-0.30in}
%\captionof{figure}{
%\vspace{-0.20in}
%\label{fig:Paris1}}
%\end{strip}

%%%%%%%%% ABSTRACT
\begin{abstract}
Neural signed distance functions (SDFs) have shown remarkable capability in representing geometry with details. However, without signed distance supervision, it is still a challenge to infer SDFs from point clouds or multi-view images using neural networks. In this paper, we claim that gradient consistency in the field, indicated by the parallelism of level sets, is the key factor affecting the inference accuracy. Hence, we propose a level set alignment loss to evaluate the parallelism of level sets, which can be minimized to achieve better gradient consistency. Our novelty lies in that we can align all level sets to the zero level set by constraining gradients at queries and their projections on the zero level set in an adaptive way. Our insight is to propagate the zero level set to everywhere in the field through consistent gradients to eliminate uncertainty in the field that is caused by the discreteness of 3D point clouds or the lack of observations from multi-view images. Our proposed loss is a general term which can be used upon different methods to infer SDFs from 3D point clouds and multi-view images. Our numerical and visual comparisons demonstrate that our loss can significantly improve the accuracy of SDFs inferred from point clouds or multi-view images under various benchmarks. Code and data are available at \url{https://github.com/mabaorui/TowardsBetterGradient}.
%\href{https://github.com/mabaorui/TowardsBetterGradient}{https://github.com/mabaorui/TowardsBetterGradient}.
\end{abstract}

%%%%%%%%% BODY TEXT

\section{Introduction}
Signed distance functions (SDFs) have shown remarkable abilities in representing high fidelity 3D geometry~\cite{DBLP:journals/corr/abs-1901-06802,Park_2019_CVPR,aminie2022,Jiang2019SDFDiffDRcvpr,Zhizhong2021icml,neuslingjie,Yu2022MonoSDF,yiqunhfSDF,DBLP:journals/corr/abs-2105-02788,takikawa2021nglod,tancik2020fourfeat,Baoruicvpr2023,chou2022gensdf,wang2022rangeudf,NeuralPoisson,VisCovolume,Peng2021SAP,huang2022neuralgalerkin,long2022neuraludf,wang2022neuris}. Current methods mainly use neural networks to learn SDFs as a mapping from 3D coordinates to signed distances. Using gradient descent, we can train neural networks by adjusting parameters to minimize errors to either signed distance ground truth~\cite{DBLP:journals/corr/abs-1901-06802,Park_2019_CVPR,aminie2022,jiang2020lig,DBLP:conf/eccv/ChabraLISSLN20} or signed distances inferred from 3D point clouds~\cite{Zhizhong2021icml,DBLP:conf/icml/GroppYHAL20,Atzmon_2020_CVPR,zhao2020signagnostic,atzmon2020sald,chaompi2022,sitzmann2019siren} or multi-view images~\cite{yariv2020multiview,yariv2021volume,geoneusfu,neuslingjie,Yu2022MonoSDF,yiqunhfSDF,Vicini2022sdf,wang2022neuris,guo2022manhattan}. However, factors like the discreteness in point clouds and the lack of observations in multi-view images result in 3D ambiguity, which makes inferring SDFs without ground truth signed distances remain a challenge.

Recent solutions~\cite{Jiang2019SDFDiffDRcvpr,Atzmon_2020_CVPR,gropp2020implicit,sitzmann2019siren,neuslingjie} impose additional constraints on gradients with respect to input coordinates. The gradients determine the rate of change of signed distances in a field, which is vital for the accuracy of SDFs. Specifically, Eikonal term~\cite{Jiang2019SDFDiffDRcvpr,Atzmon_2020_CVPR,gropp2020implicit} is widely used to learn SDFs, which constrains the norm of gradients to be one at any location in the field. This regularization ensures the networks to predict valid signed distances. NeuralPull~\cite{Zhizhong2021icml} constrains the directions of gradients to pull arbitrary queries onto the surface. One issue here is that these methods merely constrain gradients at single locations, without considering gradient consistency to their corresponding projections on different level sets. This results in inconsistent gradients in the field, indicated by level sets with poor parallelism, which drastically decreases the accuracy of inferred SDFs.

To resolve this issue, we introduce a level set alignment loss to pursue better gradient consistency for SDF inference without ground truth signed distances. Our loss is a general term which can be used to train different networks for learning SDFs from either 3D point clouds or multi-view images. Our key idea is to constrain gradients at corresponding locations on different level sets of the inferred SDF to be consistent. We achieve this by minimizing the cosine distance between the gradient of a query and the gradient of its projection on the zero level set. Minimize our loss is equivalent to aligning all level sets onto a reference, i.e. the zero level set, in a pairwise way. This enables us to propagate the zero level set to everywhere in the field, which eliminates uncertainty in the field that is caused by the discreteness of 3D point clouds or the lack of observations from multi-view images. Moreover, we introduce an adaptive weight to focus more on the gradient consistency nearer to the zero level set. We evaluate our loss upon the latest methods in surface reconstruction and multi-view 3D reconstruction under the widely used benchmarks. Our improvements over baselines justify not only our effectiveness but also the importance of gradient consistency to the inference of signed distance fields. Our contributions are listed below.

\begin{enumerate}[i)]
\item We introduce a level set alignment loss to achieve better gradient consistency for inferring SDFs without signed distance ground truth.
\item We justify the importance of gradient consistency to the accuracy of SDFs inferred from 3D point clouds and multi-view images, and show that aligning level sets together is an effective way of learning more consistent gradients for eliminating 3D ambiguity.
\item We show our superiority over the state-of-the-art methods in surface reconstruction from point clouds and multi-view 3D reconstruction under the widely used benchmarks.
\end{enumerate}

\section{Related Work}
Neural implicit representations have shown prominent performance in representing 3D geometry with details~\cite{DBLP:journals/corr/abs-1901-06802,Park_2019_CVPR,aminie2022,Jiang2019SDFDiffDRcvpr,Zhizhong2021icml,neuslingjie,Yu2022MonoSDF,yiqunhfSDF,mildenhall2020nerf,Oechsle2021ICCV,MeschederNetworks,wenxincvpr2022,Han2019ShapeCaptionerGC,chou2022gensdf}. With signed distances and occupancy labels as supervision, we can learn neural implicit representations as a regression~\cite{Park_2019_CVPR} or classification~\cite{MeschederNetworks} problem. In the following, we focus on reviewing methods inferring supervision from 3D point clouds~\cite{Jiang2019SDFDiffDRcvpr,Atzmon_2020_CVPR,gropp2020implicit} and multi-view images~\cite{mildenhall2020nerf}.

\noindent\textbf{Supervision from 3D Point Clouds. }Some methods learn SDFs or occupancy with 3D point clouds as conditions. They require signed distances and occupancy labels as supervision to learn global priors~\cite{Mi_2020_CVPR,Genova:2019:LST,jia2020learning,Liu2021MLS,tang2021sign,Peng2020ECCV,ErlerEtAl:Points2Surf:ECCV:2020,Boulch_2022_CVPR,aminie2022} or local priors~\cite{Williams_2019_CVPR,Tretschk2020PatchNets,DBLP:conf/eccv/ChabraLISSLN20,jiang2020lig,Boulch_2022_CVPR,DBLP:conf/cvpr/LiWLSH22,ChaoSparse}, which can be generalized to unseen cases. With the ground truth field, these methods get benefits including perfect scalar fields with consistent gradients, but struggle to generalize the learned priors to unseen cases with large geometry variations.

Some other methods infer SDFs without supervision by training neural networks to overfit to single point cloud. These methods require additional constraints~\cite{DBLP:conf/icml/GroppYHAL20,Atzmon_2020_CVPR,zhao2020signagnostic,atzmon2020sald,DBLP:journals/corr/abs-2106-10811,yifan2020isopoints}, specially designed operations~\cite{Peng2021SAP,Zhizhong2021icml,chaompi2022} or normals~\cite{Needle3DPoints,DBLP:conf/cvpr/WilliamsTBZ21,li2023NeAF} to estimate signed distances or occupancy using point clouds as a reference. Using similar idea, we can infer unsigned distances from point clouds~\cite{chibane2020neural,Zhou2022CAP-UDF}. Using inferred signed distances, some methods use the inferred SDFs as priors, and then guide the SDF inference from a novel point cloud~\cite{DBLP:conf/cvpr/MaLZH22,DBLP:conf/cvpr/MaLH22}.

\noindent\textbf{Supervision from Multi-View Images. } With multi-view images as supervision, classic multi-view stereo (MVS)~\cite{schoenberger2016sfm,schoenberger2016mvs} methods use multi-view consistency to estimate depth maps. With differentiable renderers~\cite{Jiang2019SDFDiffDRcvpr,Vicini2022sdf,handrwr2020}, we can render images from the learned SDFs, and refine the learned SDFs by minimizing errors between the rendered images and ground truth images. Similarly, DVR~\cite{DVRcvpr} and IDR~\cite{yariv2020multiview} infer the radiance on surfaces, where IDR models view direction as a condition to reconstruct high frequency details. However, these methods focus on surfaces, which makes them require masks of objects during optimization. Hence, we can not use them to reveal structures for scenes, where no masks are available.

NeRF~\cite{mildenhall2020nerf} and the following work~\cite{park2021nerfies,mueller2022instant,ruckert2021adop,yu_and_fridovichkeil2021plenoxels} were proposed for novel view synthesis, and render images from radiance field use volume rendering without requiring masks. By simultaneously modelling geometry and color, we can infer signed distance or occupancy fields by minimizing rendering errors. With samples on rays shooting from pixels into the field, unisurf~\cite{Oechsle2021ICCV} and NeuS~\cite{neuslingjie} use a revised rendering procedure to render occupancy and signed distance fields with radiance into pixel colors. Following methods improve accuracy of implicit functions using multi-view consistency~\cite{GEOnEUS2022,neuslingjie,Yu2022MonoSDF,yiqunhfSDF,ZhizhongSketch2020,zhizhongiccv2021finepoints} or additional priors including depth~\cite{Yu2022MonoSDF,Azinovic_2022_CVPR,Zhu2022CVPR}, normals~\cite{Yu2022MonoSDF,wang2022neuris,guo2022manhattan}.

The SDFs inferred these methods are not accurate, due to the poor gradient consistency in signed distance fields, indicated by level sets with poor parallelism. This is the key factor that impacts on the accuracy of inferred SDFs through neural rendering in a multi-view context or reasoning on point clouds. We improve gradient consistency by aligning level sets on the zero level set via minimizing a level set alignment loss. Our loss is a general term that can be used upon different methods.

\section{Method}
\noindent\textbf{Neural SDFs and Level Sets. }We focus on inferring an SDF $f$ from a 3D point cloud or a set of multi-view images which does not provide ground truth signed distances. $f$ predicts a signed distance $s\in\mathbb{R}$ for an arbitrary query point $\bm{q}\in\mathbb{R}^3$, as formulated by,

\begin{equation}
\label{eq:sdf}
s=f_{\theta}(\bm{q}),
%\vspace{-0.20in}
\end{equation}

\noindent where we use a neural network parameterized by $\theta$ to learn the SDF $f$. Level sets of $f_{\theta}$ are denoted as $\{\mathcal{S}_l\}$, each of which is a set of points where $f_{\theta}$ takes on a given constant value $l$,

\begin{equation}
\label{eq:levelset}
\mathcal{S}_l=\{\bm{q}|f_{\theta}(\bm{q})=l\},
%\vspace{-0.20in}
\end{equation}

\noindent where we regard zero level set $\mathcal{S}_0$ as the surface of the scene. We extract the surface as a triangle mesh by running the marching cubes algorithm~\cite{Lorensen87marchingcubes}.

\noindent\textbf{Inferring SDFs. }Without signed distance ground truth, current methods infer SDFs by mining supervision from 3D point clouds with normals~\cite{sitzmann2019siren,Atzmon_2020_CVPR,gropp2020implicit}, 3D point clouds without normals~\cite{Peng2021SAP,Zhizhong2021icml,chaompi2022}, or multi-view images~\cite{yariv2020multiview,yariv2021volume,geoneusfu,neuslingjie,Yu2022MonoSDF,yiqunhfSDF,Vicini2022sdf,wang2022neuris,guo2022manhattan}. Although these methods use supervision in different modalities, all of them minimize a general form of loss function $E$ to infer the SDF $f_{\theta}$ below,

\begin{equation}
\label{eq:loss}
\min_{\theta} E(T(f_{\theta}),\bm{G}),
%\vspace{-0.20in}
\end{equation}

\noindent where $\bm{G}$ is the supervision including 3D point clouds with or without normals or multi-view images, $T$ is a transformation function that transforms signed distances into a representation in the same modality of $\bm{G}$, and $E$ is the metric function that evaluates the error between the representation transformed from $f_{\theta}$ and the ground truth supervision $\bm{G}$. More specifically, NeuralPull~\cite{Zhizhong2021icml} uses 3D point clouds without normals as $\bm{G}$, then the function $T$ projects a query $\bm{q}$ on $\bm{G}$ using its singed distance $f_{\theta}(\bm{q})$ and gradient $\nabla f_{\theta}(\bm{q})$, and the loss $E$ is mean squared error (MSE) between query projections and ground truth $\bm{G}$. Siren~\cite{sitzmann2019siren} uses 3D point clouds with normals as $\bm{G}$, the function $T$ is the Eikonal term regulating gradients and MSE over signed distances of surface points, the loss $E$ is an energy based metric. NeuS~\cite{neuslingjie} uses a set of multi-view images as $\bm{G}$, then uses volume rendering as the function $T$ to render $f_{\theta}$ along with radiance into images, and compares the rendered images to $\bm{G}$ using a MSE $E$.

\begin{figure*}[tb]
  \centering
  % the following command controls the width of the embedded PS file
  % (relative to the width of the current column)
  %\includegraphics[width=.95\linewidth, bb=39 696 126 756]{figures/definition3.eps}
   \includegraphics[width=\linewidth]{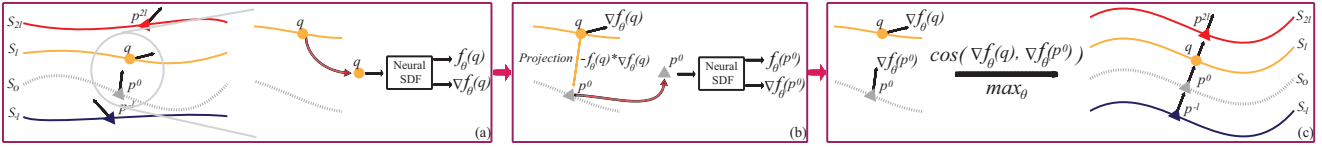}
  % replacing the above command with the one below will explicitly set
  % the bounding box of the PS figure to the rectangle (xl,yl),(xh,yh).
  % It will also prevent LaTeX from reading the PS file to determine
  % the bounding box (i.e., it will speed up the compilation process)
  % \includegraphics[width=.95\linewidth, bb=39 696 126 756]{sampleFig}
  %
  %
  \vspace{-0.2in}
\caption{\label{fig:overview}Overview of our level set alignment loss. We minimize our loss to pursue better gradient consistency in (c). The inconsistent gradient at a query $\bm{q}$ in (a) and its projections on zero level set in (b) are constrained to be consistent.}
%\vspace{-0.1in}
\end{figure*}

\noindent\textbf{Gradient Consistency. }Our main contribution lies in pursuing better gradient consistency. We illustrate gradient consistency in the field using one query $\bm{q}$ in Fig.~\ref{fig:overview}. If gradients are consistent, as shown in Fig.~\ref{fig:overview} (c), the gradient at query $\bm{q}$ and the gradient at its projection on each level set $\mathcal{S}_l$ should point to the same direction, which leads to level sets with great parallelism, while inconsistent gradients shown in Fig.~\ref{fig:overview} (a) result in level sets with poor parallelism. To evaluate gradient consistency at a query $\bm{q}$, we use cosine distance between gradients at query and its projection on a level set $\mathcal{S}_l$,

\begin{equation}
\label{eq:consistency}
c(\bm{q},\mathcal{S}_l)=1-\frac{\nabla f_{\theta}(\bm{q})\cdot\nabla f_{\theta}(\bm{p}^l)}{||\nabla f_{\theta}(\bm{q})||_2\cdot||\nabla f_{\theta}(\bm{p}^l)||_2},
%\vspace{-0.20in}
\end{equation}

\noindent where $\bm{p}^l$ is the projection of $\bm{q}$ on the level set $\mathcal{S}_l$. $c(\bm{q},\mathcal{S}_l)$ is in a range of $[0,2]$, where 0 indicates $f_{\theta}(\bm{q})$ and $f_{\theta}(\bm{p}^l)$ are pointing to the same direction, which are the most consistent.

\begin{figure*}[tb]
  \centering
  % the following command controls the width of the embedded PS file
  % (relative to the width of the current column)
  %\includegraphics[width=.95\linewidth, bb=39 696 126 756]{figures/definition3.eps}
   \includegraphics[width=\linewidth]{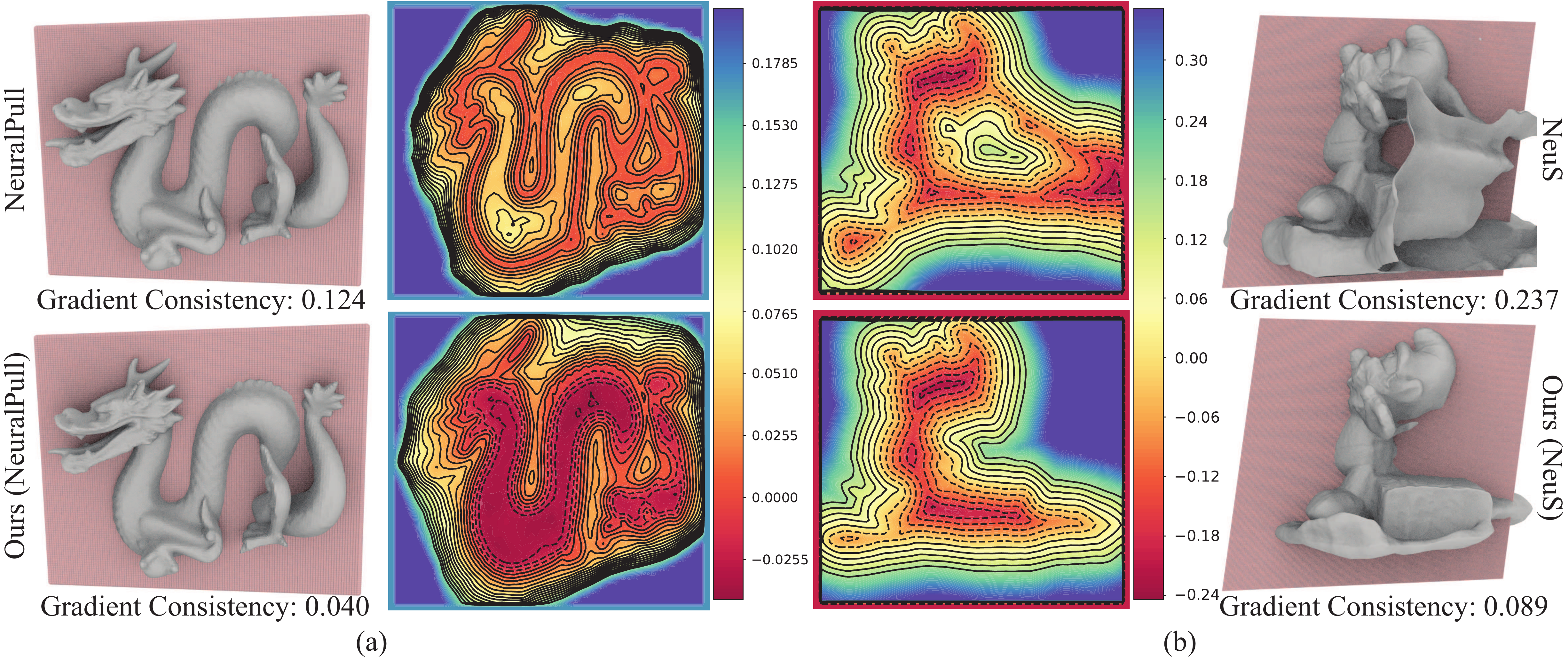}
  % replacing the above command with the one below will explicitly set
  % the bounding box of the PS figure to the rectangle (xl,yl),(xh,yh).
  % It will also prevent LaTeX from reading the PS file to determine
  % the bounding box (i.e., it will speed up the compilation process)
  % \includegraphics[width=.95\linewidth, bb=39 696 126 756]{sampleFig}
  %
  %
  \vspace{-0.25in}
\caption{\label{fig:LevelSets}Visualization of level sets on a cross section. We pursue better gradient consistency in a field learned from 3D point clouds in (a) and multi-view images in (b). We minimize our level set alignment loss with NeuralPull in (a) and NeuS in (b), which leads to more accurate SDFs with better parallelism of level sets and less artifacts in empty space.}
\vspace{-0.1in}
\end{figure*}

One issue here is that it costs extensive computation if we evaluate $c(\bm{q},\mathcal{S}_l)$ on each level set $\mathcal{S}_l$. Our solution here is to use zero level set $\mathcal{S}_0$ as a reference and project all queries onto the reference to evaluate the gradient consistency. In this pairwise way, we associate all level sets $\mathcal{S}_l$ to the zero level set $\mathcal{S}_0$, which can propagate consistency to all level sets through the projections on $\mathcal{S}_0$ since we randomly sample a large amount of queries in each iteration during optimization. Hence, we pursue a better gradient consistency by minimizing Eq.~\ref{eq:consistency} over all sampled queries $\mathcal{Q}$,

\begin{equation}
\label{eq:consistencyloss}
\min_{\theta}\sum_{\bm{q}\in\mathcal{Q}}c(\bm{q},\mathcal{S}_0).
%\vspace{-0.20in}
\end{equation}

Minimizing the loss in Eq.~\ref{eq:consistencyloss} is equivalent to align all level sets to the zero level set, which we named it as level set alignment loss, as illustrated in Fig.~\ref{fig:overview} (c).

\noindent\textbf{Loss Function. }We use our level set alignment loss upon methods for inferring neural SDF without signed distance ground truth. We formulate our loss function by combining Eq.~\ref{eq:loss} and Eq.~\ref{eq:consistencyloss} below,

\begin{equation}
\label{eq:totalloss}
\min_{\theta} E_{\bm{q}\in\mathcal{Q}}(T(f_{\theta}(\bm{q})),\bm{G})+\alpha\sum_{\bm{q}\in\mathcal{Q}}\beta_{\bm{q}}c(\bm{q},\mathcal{S}_0).
%\vspace{-0.20in}
\end{equation}

\noindent where $\alpha$ is the balance weight for our level set alignment loss, and it scales the per point weight $\beta_{\bm{q}}$ which is an adaptive weight indicating the importance of each query $\bm{q}$, as defined below,

\begin{equation}
\label{eq:totallossweight}
\beta_{\bm{q}}=exp(-\delta*|f_{\theta}(\bm{q})|),
%\vspace{-0.20in}
\end{equation}

\noindent where we model $\beta_{\bm{q}}$ according to the predicted signed distance, which aims to encourage the optimization to focus more on the area near the surface. Another option to replace $|f_{\theta}(\bm{q})|$ is to use the distance between $\bm{q}$ and its nearest point in point cloud representing a surface. However, finding the nearest point for each query $\bm{q}$ increases computational burden. Moreover, we can not use it in scenarios without point clouds such as multi-view images.

By optimizing the objective function in Eq.~\ref{eq:totalloss}, we can achieve more consistent gradients in the field, as illustrated in Fig.~\ref{fig:LevelSets}. By optimizing with our level set alignment loss, we reformulate the loss function of NeuralPull~\cite{Zhizhong2021icml} into Eq.~\ref{eq:totalloss}, which improves the gradient consistency in the field learned from 3D point clouds without normals. Fig.~\ref{fig:LevelSets} (a) shows that we improve the parallelism of level sets, especially near the surface and inside of the dragon, where we visualize the signed distance field on a cross section of the reconstructed surface. This enables us to eliminate the swollen effect on the reconstructed surface of NeuralPull, which achieves a more compact surface with sharper edges. Similarly, we reformulate the loss function of NeuS~\cite{neuslingjie} into Eq.~\ref{eq:totalloss} by adding our level set alignment loss, which improves the gradient consistency in the field learned from multi-view images. Fig.~\ref{fig:LevelSets} (b) shows that the better gradient consistency leads to level sets with better parallelism, which propagates the zero level sets to everywhere in the field. This is a key factor to eliminate the artifacts in the empty space.

\noindent\textbf{Projections on the Zero Level Set. }We project a query $\bm{q}$ onto the zero level set $\mathcal{S}_0$, and use the projection $\bm{p}^0$ to evaluate the gradient consistency defined in Eq.~\ref{eq:consistency}. As illustrated in Fig.~\ref{fig:overview}(b), we follow the differentiable pulling operation in~\cite{Zhizhong2021icml}, and use the predicted signed distance $f_{\theta}(\bm{q})$ and the gradient $\nabla f_{\theta}(\bm{q})$ to project $\bm{q}$, as formulated by,

\begin{equation}
\label{eq:totallosspull}
\bm{p}^0=\bm{q}-|f_{\theta}(\bm{q})|\frac{\nabla f_{\theta}(\bm{q})}{||\nabla f_{\theta}(\bm{q})||_2}.
%\vspace{-0.20in}
\end{equation}

By replacing $\bm{p}^l$ in Eq.~\ref{eq:consistency} into $\bm{p}^0$, we obtain $c(\bm{q},\mathcal{S}_0)$ in Eq.~\ref{eq:consistencyloss} and Eq.~\ref{eq:totalloss} below,

\begin{equation}
\label{eq:consistency0}
c(\bm{q},\mathcal{S}_0)=1-\frac{\nabla f_{\theta}(\bm{q})\cdot\nabla f_{\theta}(\bm{p}^0)}{||\nabla f_{\theta}(\bm{q})||_2\cdot||\nabla f_{\theta}(\bm{p}^0)||_2}.
%\vspace{-0.20in}
\end{equation}

%\noindent where we use chain rule to further decompose the term $\nabla f_{\theta}(\bm{p}^0)$ into,
%
%\begin{equation}
%\label{eq:consistency1}
%\begin{aligned}
%&\nabla f_{\theta}(\bm{p}^0)=\frac{\partial f_{\theta}(\bm{p}^0)}{\partial \bm{p}^0}\frac{\partial \bm{p}^0}{\partial \bm{q}}\\
%&=(1-f_{\theta}(\bm{q})\nabla^2f_{\theta}(\bm{q})-(\nabla f_{\theta}(\bm{q}))^2)\frac{\partial f_{\theta}(\bm{p}^0)}{\partial \bm{p}^0},
%\end{aligned}
%%\vspace{-0.20in}
%\end{equation}
%
%
%\noindent where $\nabla^2f_{\theta}(\bm{q})$ is the second order partial derivative with respect to query $\bm{q}$. We involve $\nabla^2f_{\theta}(\bm{q})$ to constrain the change of gradients, which achieves smoother level sets.

\section{Experiments}
We conduct experiments to evaluate our method in learning neural signed distance functions for 3D reconstruction from 3D point clouds and multi-view images. We use our level set alignment loss upon different methods to improve the performance by encouraging more consistent gradients in the field. We extract the zero level set of the learned signed distance functions using the marching cubes algorithm~\cite{Lorensen87marchingcubes} as a surface. Note that we do not evaluate our performance with methods learning from signed distance ground truth, since the supervision provides perfect gradient consistency in the field, which dose not highlight our inference capability.

\subsection{Surface Reconstruction from 3D Point Clouds}
\noindent\textbf{Datasets.}We evaluate our performance under three datasets including the one released by SIREN~\cite{sitzmann2019siren}, Stanford Scanning~\cite{Curless1996AVM} and 3D Scene~\cite{DBLP:journals/tog/ZhouK13}. These datasets contains challenging cases including single objects and scenes with arbitrary topology and complex geometry. We use the point clouds in the dataset released by SIREN, which each scene contains millions points, and we sample $2$ million points for each shape or scene in Stanford Scanning and 3D Scene.

\noindent\textbf{Metrics. }We evaluate the accuracy of the learned SDFs using the error between the reconstructed meshes and ground truth. We use L1 Chamfer distance (CD) and normal consistency (NC) to measure the error. We sample $100k$ points on the reconstructed mesh and ground truth to calculate CD in Stanford Scanning dataset, and sample $1$ million points to calculate CD in SIREN dataset and 3D Scene. We also use the normals estimated on the reconstructed meshes for the calculation of NC.

\noindent\textbf{Baselines. }We report our performance with the latest methods learning SDFs from 3D point clouds including IGR~\cite{DBLP:conf/icml/GroppYHAL20}, SIREN~\cite{sitzmann2019siren}, NeuralPull~\cite{Zhizhong2021icml}. These methods infer SDFs by training neural networks to overfit single 3D point cloud, with learning priors from large scale dataset. Specifically, IGR and SIREN adopt similar strategy to infer SDFs. They use Eikonal term to constrain the length of gradients to be one at everywhere, employ additional point normals to constrain gradients at points on surface, and set signed distances on surface to be zero. While NeuralPull constrains the gradients and signed distances together via pulling a query onto the surface without normals.

\noindent\textbf{Details. }To highlight our capability of inferring consistent gradients, we do not use the ground truth normal to produce our results with IGR and SIREN, since the ground truth normal is a direct supervision for gradients. We report our result upon the baselines using their official code. We use the loss function of the baseline to replace the first term in Eq.~\ref{eq:totalloss}, which is combined with our level set alignment loss into a loss function we use to report our results. We set the weight $\alpha$ to make our loss contribute equally as the loss of the baseline.

\noindent\textbf{Comparison. }We report numerical evaluations in SIREN dataset in Tab.~\ref{table:SIRENDATA}. In these point clouds with high frequency details, SIREN performs not well without using normals as supervision. Since ground truth normals determines the distance field near point clouds, which is the key to reconstruct accurate surface. But, without normals, the other loss terms in SIREN, such as the Eikonal term, can not infer accurate signed distances. While minimizing our loss can achieve more accurate signed distance field, which reveals surfaces with details even without using normal supervision. We also report the comparison with normal supervision in our supplementary materials.

\begin{figure}[tb]
  \centering
  % the following command controls the width of the embedded PS file
  % (relative to the width of the current column)
  %\includegraphics[width=.95\linewidth, bb=39 696 126 756]{figures/definition3.eps}
   \includegraphics[width=\linewidth]{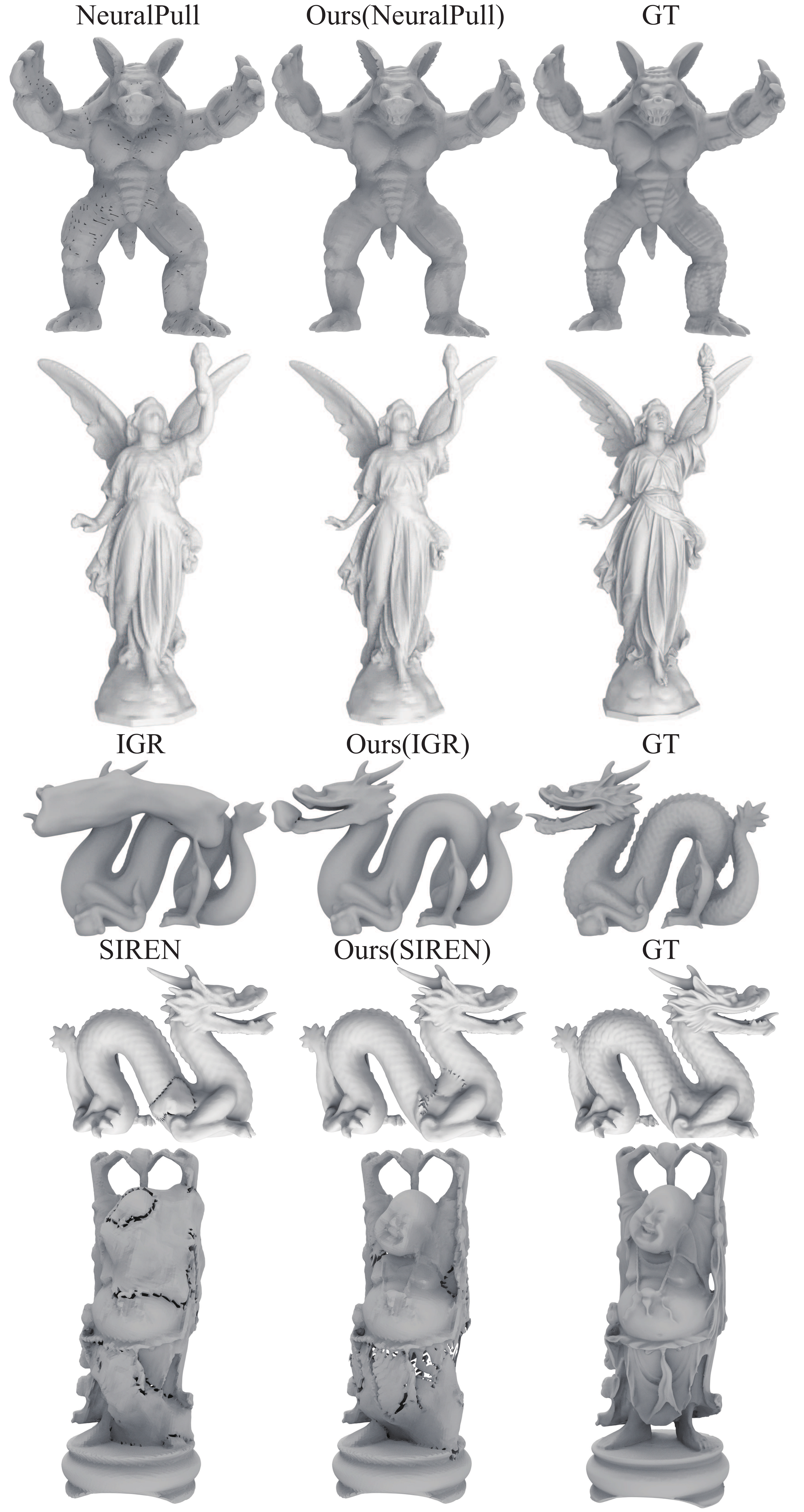}
  % replacing the above command with the one below will explicitly set
  % the bounding box of the PS figure to the rectangle (xl,yl),(xh,yh).
  % It will also prevent LaTeX from reading the PS file to determine
  % the bounding box (i.e., it will speed up the compilation process)
  % \includegraphics[width=.95\linewidth, bb=39 696 126 756]{sampleFig}
  %
  %
\caption{\label{fig:stanford}Visual comparison with baselines in 3D scene dataset.
}
\vspace{-0.05in}
\end{figure}

We report our evaluation in Stanford scanning dataset in Tab.~\ref{table:Stanford}. Without using normals as supervision, SIREN and IGR reconstruct surfaces with artifacts. With our loss, we improve the gradient consistency in the field, especially near the surfaces. As shown in Fig.~\ref{fig:stanford}, we can eliminate the artifacts in the empty area, and obtain more completed and smoother surface. Since NeuralPull can not infer the zero level set very accurately, its reconstructed surfaces look a little bit ``fat''. The swollen effect is mainly caused by inference uncertainty near surfaces. Training with our loss can make NeuralPull infer more accurate distance fields with much less uncertainty. This leads to more compact surfaces with more details, as shown in Fig.~\ref{fig:stanford}.

\begin{figure}[tb]
  \centering
  % the following command controls the width of the embedded PS file
  % (relative to the width of the current column)
  %\includegraphics[width=.95\linewidth, bb=39 696 126 756]{figures/definition3.eps}
   \includegraphics[width=\linewidth]{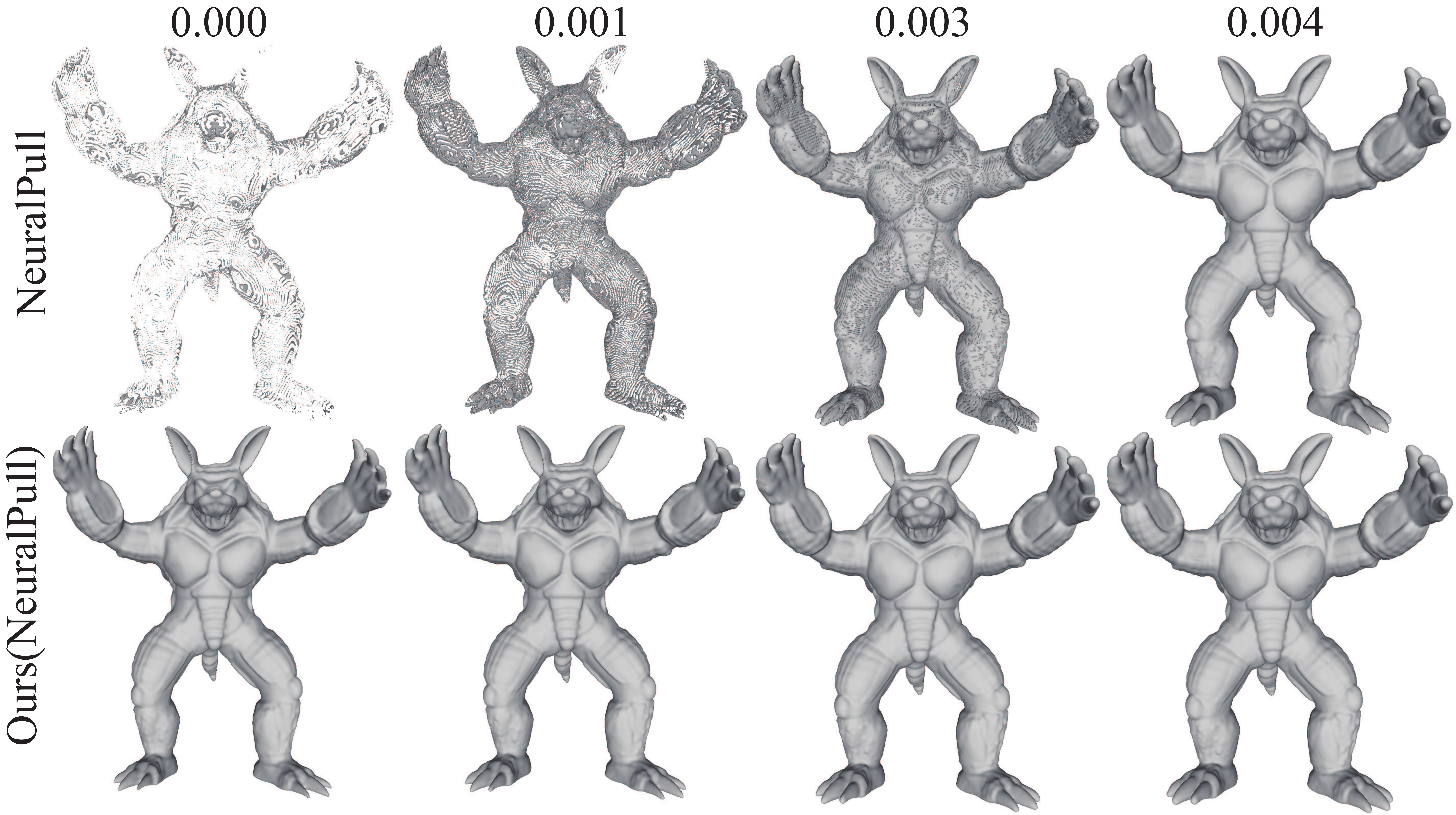}
  % replacing the above command with the one below will explicitly set
  % the bounding box of the PS figure to the rectangle (xl,yl),(xh,yh).
  % It will also prevent LaTeX from reading the PS file to determine
  % the bounding box (i.e., it will speed up the compilation process)
  % \includegraphics[width=.95\linewidth, bb=39 696 126 756]{sampleFig}
  %
  %
  \vspace{-0.2in}
\caption{\label{fig:LevelSets3D}Visual comparisons of level sets with NeuralPull.}
\vspace{-0.1in}
\end{figure}

\begin{figure}[tb]
  \centering
  % the following command controls the width of the embedded PS file
  % (relative to the width of the current column)
  %\includegraphics[width=.95\linewidth, bb=39 696 126 756]{figures/definition3.eps}
   \includegraphics[width=\linewidth]{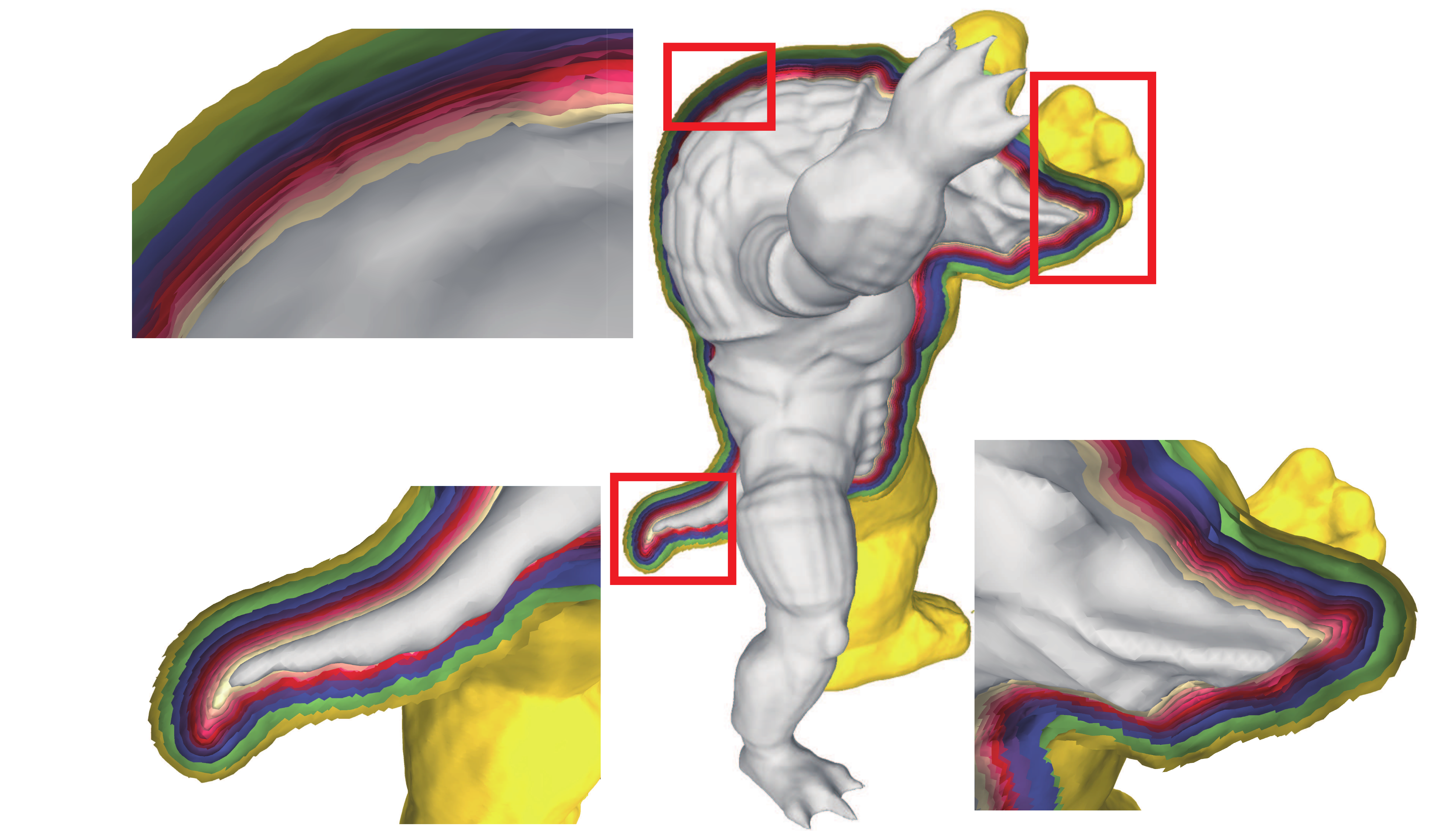}
  % replacing the above command with the one below will explicitly set
  % the bounding box of the PS figure to the rectangle (xl,yl),(xh,yh).
  % It will also prevent LaTeX from reading the PS file to determine
  % the bounding box (i.e., it will speed up the compilation process)
  % \includegraphics[width=.95\linewidth, bb=39 696 126 756]{sampleFig}
  %
  %
  \vspace{-0.2in}
\caption{\label{fig:LevelSets3D1}Visualization of level sets learned with our loss.We visualize the internal and external level sets, where the red surface represents the zero level set.}
%\vspace{-0.3in}
\end{figure}

We visualize the level sets learned with our loss in Fig.~\ref{fig:LevelSets3D} and Fig.~\ref{fig:LevelSets3D1}. The comparisons of level sets shown in Fig.~\ref{fig:LevelSets3D} indicate that better gradient consistency can achieve more completed level sets near the surface. More visualization of level sets can be found in Fig.~\ref{fig:LevelSets3D1}.

\begin{figure*}[tb]
  \centering
  % the following command controls the width of the embedded PS file
  % (relative to the width of the current column)
  %\includegraphics[width=.95\linewidth, bb=39 696 126 756]{figures/definition3.eps}
   \includegraphics[width=\linewidth]{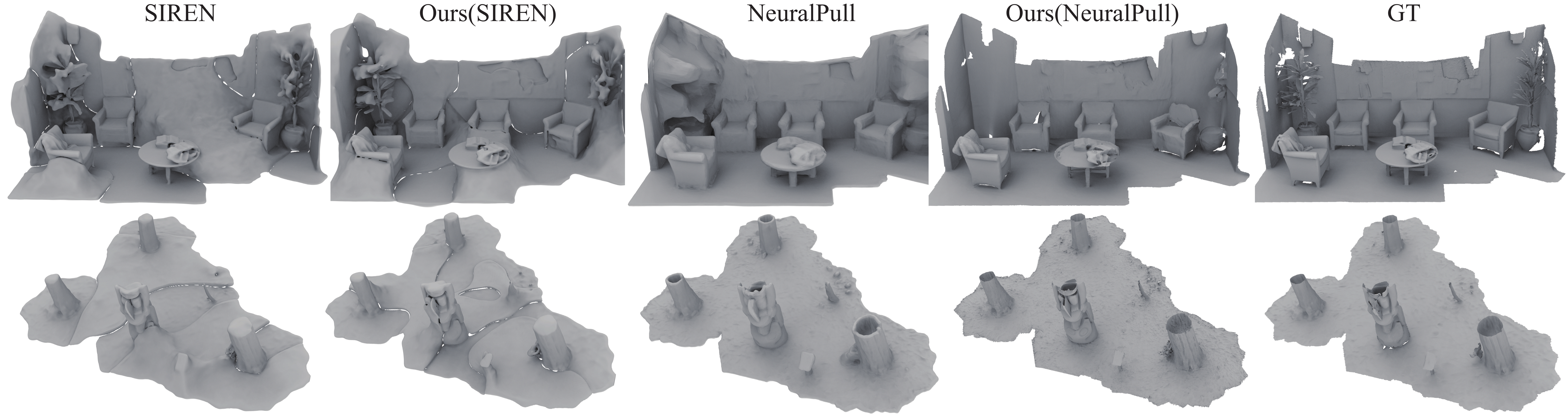}
  % replacing the above command with the one below will explicitly set
  % the bounding box of the PS figure to the rectangle (xl,yl),(xh,yh).
  % It will also prevent LaTeX from reading the PS file to determine
  % the bounding box (i.e., it will speed up the compilation process)
  % \includegraphics[width=.95\linewidth, bb=39 696 126 756]{sampleFig}
  %
  %
  \vspace{-0.2in}
\caption{\label{fig:3DScene}Visual comparison with baselines in 3D scene dataset.
}
\vspace{-0.1in}
\end{figure*}

We further evaluate our method in 3D scene dataset in Tab.~\ref{table:scenenet}. Our level set alignment loss significantly improves the performance of baselines. Visual comparisons in Fig.~\ref{fig:3DScene} illustrate that we improve the field by removing artifacts near the surface, reconstructing thinner and more compact surfaces, and sharpening surface edges. Moreover, we also report visual comparisons with methods using learned priors in our supplementary materials.

\begin{table}
\centering
\resizebox{\linewidth}{!}{
    \begin{tabular}{c|c|c||c|c}
     \hline
    &\multicolumn{2}{c||}{Thai}&\multicolumn{2}{c}{Room}\\
    \hline
    Metric&SIREN&Ours(SIREN)&SIREN&Ours(SIREN)\\
    CD&0.0043&\textbf{0.0011}&0.0189&\textbf{0.0023}\\
   \hline
   \end{tabular}}
   %\vspace{-0.1in}
   \caption{Numerical comparison in SIREN dataset.}
   \label{table:SIRENDATA}
\end{table}

\begin{table}
\centering
\resizebox{\linewidth}{!}{
    \begin{tabular}{c|c|c||c|c||c|c}
     \hline
    &SIREN&Ours(SIREN)&IGR&Ours(IGR)&NP&Ours(NP)\\
     \hline
    CD&0.0130&\textbf{0.0129}&0.020&\textbf{0.011}&0.006&\textbf{0.004}\\
    \hline
    NC&0.942&\textbf{0.948}&0.946&\textbf{0.947}&0.955&\textbf{0.958}\\
   \hline
   \end{tabular}}
   %\vspace{-0.1in}
   \caption{Numerical comparison in Stanford scanning.}
   \vspace{-0.1in}
   \label{table:Stanford}
\end{table}

\begin{table*}[!]
\centering
%\resizebox{\linewidth}{!}{
    \begin{tabular}{c|c|c||c|c||c|c||c|c||c|c}
     \hline
        &\multicolumn{2}{c||}{Burghers}&\multicolumn{2}{c||}{Lounge}&\multicolumn{2}{c||}{Copyroom}&\multicolumn{2}{c||}{Stonewall}&\multicolumn{2}{c}{Totempole}\\
        \hline
        &CD&NC&CD&NC&CD&NC&CD&NC&CD&NC\\
     \hline
     MPU~\cite{OhtakeBATS03}&0.456&0.720&0.206&0.817&0.062&0.832&0.428&0.800&0.671&0.763\\
     ConvOcc~\cite{Peng2020ECCV}&0.077&0.865&0.042&0.857&0.045&0.848&0.066&0.866&0.016&0.925\\
     LIG~\cite{jiang2020lig}&0.018&0.904&0.017&0.910&0.018&0.910&0.020&0.928&0.023&0.917\\
     \hline\hline
     NP~\cite{Zhizhong2021icml}&0.010&0.883&0.059&0.857&0.011&0.884&0.007&0.868&0.010&0.765\\
     Ours (NP)&\textbf{0.008}&\textbf{0.947}&\textbf{0.020}&\textbf{0.936}&\textbf{0.009}&\textbf{0.941}&\textbf{0.006}&\textbf{0.972}&\textbf{0.008}&\textbf{0.968}\\
     \hline\hline
     SIREN~\cite{sitzmann2019siren}&0.025&0.944&0.064&0.933&0.032&0.917&\textbf{0.026}&0.938&0.032&\textbf{0.952}\\
     Ours (SIREN)&\textbf{0.016}&\textbf{0.948}&\textbf{0.021}&\textbf{0.929}&\textbf{0.026}&\textbf{0.922}&0.031&\textbf{0.940}&\textbf{0.028}&0.937\\
   \hline
   \end{tabular}%}
   \vspace{-0.05in}
   \caption{Numerical comparison with baselines in 3D scene dataset.}
   %\vspace{-0.2in}
   \label{table:scenenet}
\end{table*}

\subsection{3D Reconstruction from Multi-view Images}
\noindent\textbf{Dataset. }We further evaluate our loss in reconstructing 3D shapes from multi-view images in the DTU datatset~\cite{jensen2014large}. Following previous methods~\cite{Oechsle2021ICCV,yariv2020multiview,yariv2021volume,geoneusfu,neuslingjie,Yu2022MonoSDF,yiqunhfSDF}, we report our results on the widely used $15$ scenes, each of which shows single object with background in $49$ to $64$ images with different shape appearances. For larger scale scenes, we report our results under ScanNet~\cite{dai2017bundlefusion}. For fair comparison, we follow MonoSDF~\cite{Yu2022MonoSDF} to conduct evaluations using the same scenes.

\noindent\textbf{Metrics. }For evaluations under DTU dataset, we use L1 Chamfer distance to evaluate the error of points randomly sampled on the reconstructed surfaces compared to the ground truth. Following previous methods~\cite{Oechsle2021ICCV,yariv2020multiview,yariv2021volume,geoneusfu,neuslingjie,Yu2022MonoSDF,yiqunhfSDF}, we clean the reconstructed meshes using the respective masks. We use the official evaluation code released by the DTU dataset to measure our accuracy. For evaluations under ScanNet~\cite{dai2017bundlefusion}, we use the same metrics as MonoSDF, which includes Chamfer distance, F-Scores with a threshold of 5cm, and normal consistency to measure the error between the reconstructed surface and the ground truth surface in ScanNet.

\noindent\textbf{Baselines. }We add our loss on the latest methods for learning SDFs from multi-view images. we use NeuS~\cite{neuslingjie} and MonoSDF~\cite{Yu2022MonoSDF} as baselines. NeuS does not use priors, and infer an SDF using multi-view consistency through volume rendering. MonoSDF adopts the same strategy and learns SDFs with depth and normal priors on images.

\noindent\textbf{Details. }We use the official code released by NeuS and MonoSDF to produce our results with our level set alignment loss. We use the loss function of the baseline to replace the first term in Eq.~\ref{eq:totalloss}, which is combined with our level set alignment loss into a loss function we use to report our results. We set the weight $\alpha$ to make our loss contribute equally as the loss of the baseline.

\begin{figure}[tb]
  \centering
  % the following command controls the width of the embedded PS file
  % (relative to the width of the current column)
  %\includegraphics[width=.95\linewidth, bb=39 696 126 756]{figures/definition3.eps}
   \includegraphics[width=\linewidth]{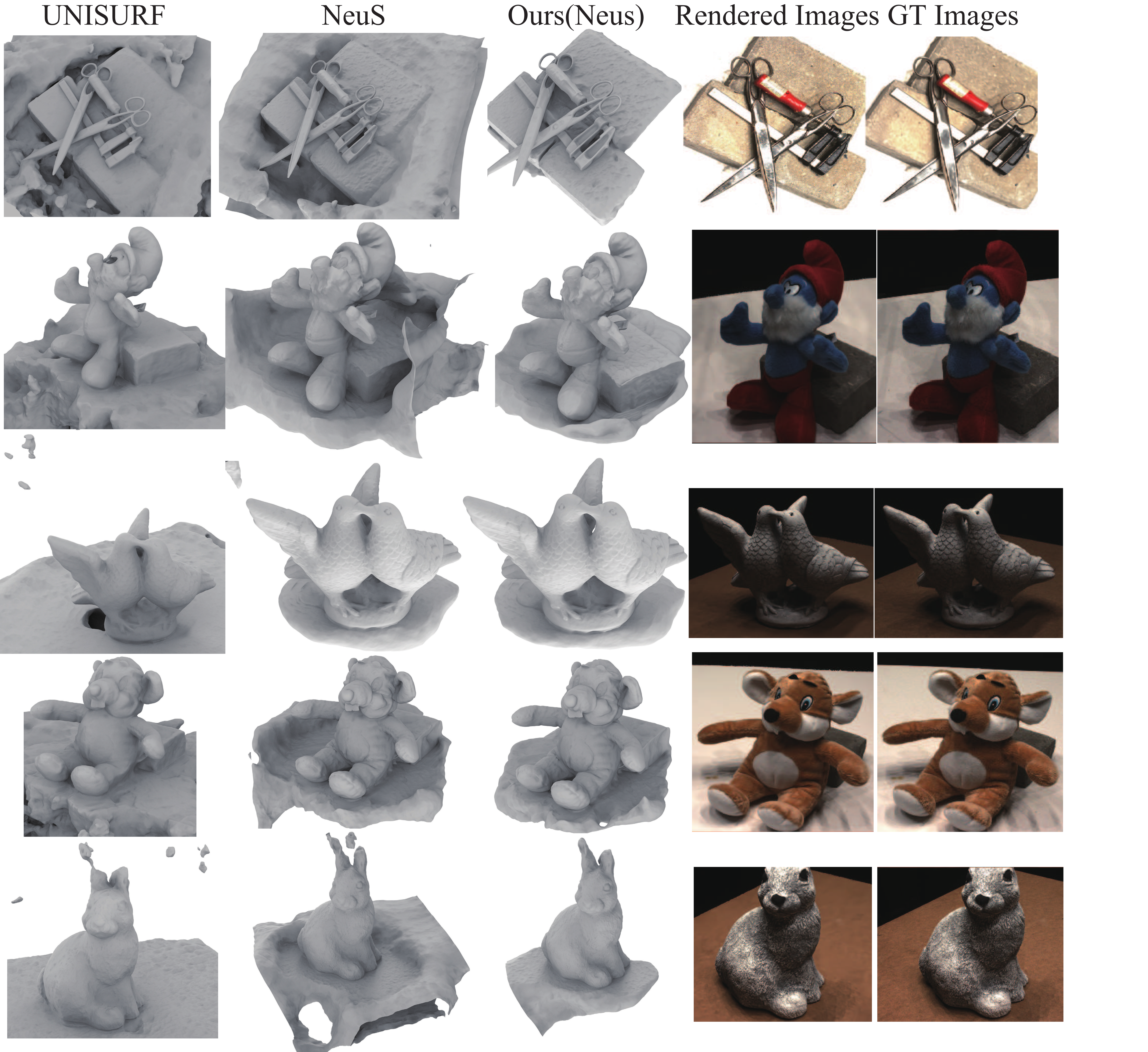}
  % replacing the above command with the one below will explicitly set
  % the bounding box of the PS figure to the rectangle (xl,yl),(xh,yh).
  % It will also prevent LaTeX from reading the PS file to determine
  % the bounding box (i.e., it will speed up the compilation process)
  % \includegraphics[width=.95\linewidth, bb=39 696 126 756]{sampleFig}
  %
  %
\caption{\label{fig:dtu}Visual comparison with baselines in DTU dataset.
}
\vspace{-0.1in}
\end{figure}

\noindent\textbf{Comparison. }We report numerical evaluations in DTU Tab.~\ref{table:DTU}. We achieve better performance in 10 out of 15 scenes, and get comparable results in the other 5 scenes. In terms of the Chamfer distance, our improvements over NeuS are subtle. The reason is that the advantages of better gradient consistency lie in the ability of improving the smoothness of surfaces and removing artifacts in empty space. However, the smoothness does not significantly improve the numerical results, and artifacts in empty space has been cleaned using respective masks following the evaluation protocol. Hence, we highlight our improvements in visual comparison in Fig.~\ref{fig:dtu}, where we show the reconstructed surfaces before the cleaning. More analysis and comparisons with NeuS can be found in supplementary materials.

As we can see, unisurf and NeuS learn neural implicit fields with lots of uncertainty, which is caused by the lack of multi-view consistency constraints or the ambiguity with the textureless background. This uncertainty results in artifacts especially in empty space. By minimizing our level set alignment loss, we can propagate the zero level set to all other level sets everywhere in the field through consistent gradients, which eliminates the uncertainty that can not get inferred from multi-view images. Hence, our results produce much less artifacts even in the area that few images can cover. We also show our rendered images as reference.

\begin{table*}[!]
\centering
\resizebox{\linewidth}{!}{
    \begin{tabular}{cccccccccccccccc|c}
    %&\multicolumn{16}{c}{Scene ID}\\
    \hline
    Method&24&37&40&55&63&65&69&83&97&105&106&110&114&118&122&Mean\\
    \hline
    COLMAP~\cite{schoenberger2016mvs}&$\overline{0.81}$&2.05&0.73&1.22&1.79&1.58&1.02&3.05&1.40&2.05&1.00&1.32&0.49&0.78&1.17&1.36\\
    NeRF~\cite{mildenhall2020nerf}&1.90&1.60&1.85&0.58&2.28&1.27&1.47&1.67&2.05&1.07&0.88&2.53&1.06&1.15&0.96&1.49\\
    UNISURF~\cite{Oechsle2021ICCV}&1.32&1.36&1.72&0.44&1.35&0.79&0.80&1.49&1.37&0.89&0.59&1.47&0.46&0.59&0.62&1.02\\
    VolSDF~\cite{yariv2021volume}&1.14&1.26&0.81&0.49&1.25&0.70&0.72&1.29&1.18&$\overline{0.70}$&0.66&$\overline{1.08}$&0.42&0.61&0.55&0.86\\
    \hline
    NeuS~\cite{neuslingjie}&1.37&1.21&\textbf{0.73}&\textbf{0.40}&1.20&0.70&0.72&\textbf{1.01}&1.16&\textbf{0.82}&0.66&1.69&0.39&0.49&\textbf{0.51}&0.87\\
    Ours(NeuS)&\textbf{0.88}&\textbf{0.90}&0.80&0.41&\textbf{1.13}&\textbf{0.63}&\textbf{0.58}&1.37&\textbf{1.157}&0.83&\textbf{0.51}&\textbf{1.26}&\textbf{0.33}&\textbf{0.48}&0.52&\textbf{0.78}\\
    \hline
    \end{tabular}
}
\vspace{-0.1in}
\caption{Numerical comparison with baselines in DTU dataset. The bars above numbers indicate the best.}
\vspace{-0.1in}
\label{table:DTU}
\end{table*}

%\begin{table}
%\centering
%%\resizebox{\linewidth}{!}{
%    \begin{tabular}{c|c|c}
%     \hline
%    Method&CD&F-score\\
%    \hline
%    MonoSDF&0.042&0.733\\
%    Ours (MonoSDF)&\textbf{0.041}&\textbf{0.750}\\
%   \hline
%   \end{tabular}%}
%   \caption{ScanNet}
%   \label{table:ScanNet}
%\end{table}

\begin{table*}
\centering
\resizebox{\linewidth}{!}{
    \begin{tabular}{c|c|c|c|c|c||c|c}
     \hline
    &COLMAP~\cite{schoenberger2016mvs}&UNISURF~\cite{Oechsle2021ICCV}&NeuS~\cite{neuslingjie}&VolSDF~\cite{yariv2021volume}&Manhattan-SDF~\cite{guo2022manhattan}&MonoSDF~\cite{Yu2022MonoSDF}&Ours(MonoSDF)\\
    \hline
    CD&0.141&0.359&0.194&0.267&0.070&0.042&\textbf{0.041}\\
    F-score&0.537&0.267&0.291&0.364&0.602&0.733&\textbf{0.750}\\
   \hline
   \end{tabular}}
   \vspace{-0.1in}
   \caption{Numerical comparison with the state-of-the-art in ScanNet. We show normals as the color map on the surface.}
   \vspace{-0.1in}
   \label{table:ScanNet}
\end{table*}

\begin{figure}[tb]
  \centering
  % the following command controls the width of the embedded PS file
  % (relative to the width of the current column)
  %\includegraphics[width=.95\linewidth, bb=39 696 126 756]{figures/definition3.eps}
   \includegraphics[width=\linewidth]{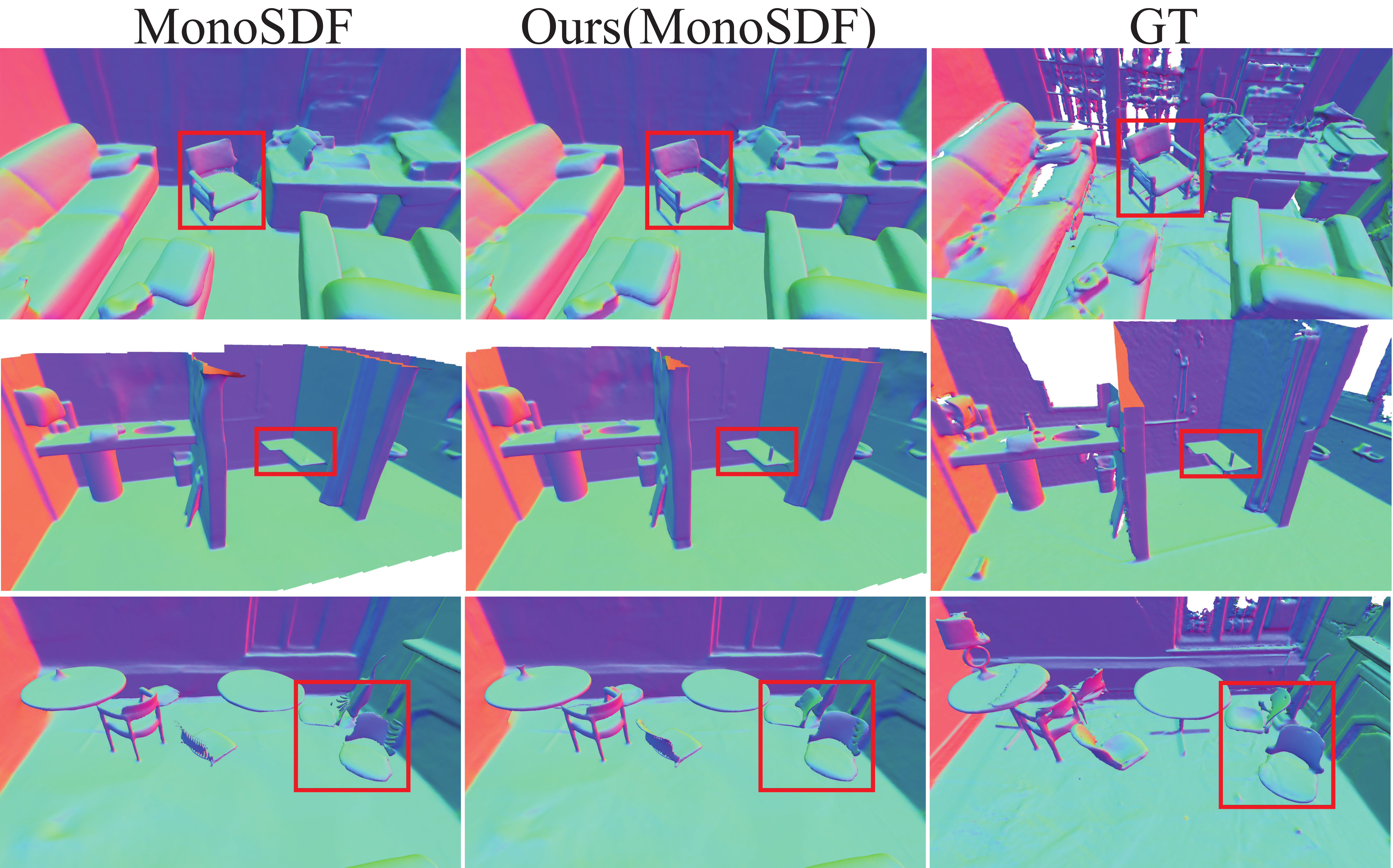}
  % replacing the above command with the one below will explicitly set
  % the bounding box of the PS figure to the rectangle (xl,yl),(xh,yh).
  % It will also prevent LaTeX from reading the PS file to determine
  % the bounding box (i.e., it will speed up the compilation process)
  % \includegraphics[width=.95\linewidth, bb=39 696 126 756]{sampleFig}
  %
  %
  \vspace{-0.2in}
\caption{\label{fig:scannet}Visual comparison with baselines in ScanNet.
}
\vspace{-0.1in}
\end{figure}

We further evaluate our method in ScanNet. We report average performance over each scene in Tab.~\ref{table:ScanNet}. The numerical comparison show that we achieve the best performance among the state-of-the-art methods. Visual comparisons in Fig.~\ref{fig:scannet} show that better gradient consistency reveals more geometry details.

\subsection{Ablation Studies}
We conduct ablation studies to justify the effectiveness of modules in our method. We use NeuralPull as a baseline and train it using our loss as one term in the loss function. We report our ablation studies under 3D scene dataset.

\noindent\textbf{Weights. }We explore the effect of our loss by adjusting the weight $\alpha$ in Eq.~\ref{eq:totalloss}. We report our results with different candidates $\{0,0.001,0.01,0.1,1.0\}$. The comparison in Tab.~\ref{table:Weight} shows that our level set alignment loss can improve the accuracy of inferred SDFs, it may affect the optimization to converge if we weight it too much.

\begin{table}
\centering
\resizebox{\linewidth}{!}{
    \begin{tabular}{c|c|c|c|c|c}
     \hline
    Weight $\alpha$&0&0.001&0.01&0.1&1.0\\
    \hline
    CD$\times 100$&0.586&0.550&\textbf{0.394}&0.433&0.601\\
    NC&0.955&0.955&\textbf{0.958}&0.957&0.940\\
   \hline
   \end{tabular}}
   \vspace{-0.1in}
   \caption{Effect of weight $\alpha$.}
   \label{table:Weight}
\end{table}

\noindent\textbf{Adaptive per Point Weights. }We show the effect of the adaptive weight $\beta_{\bm{q}}$ for each query in Tab.~\ref{table:AdaptiveWeight}. We report the result without the weight $\beta_{\bm{q}}$, and the result with $\beta_{\bm{q}}$ that is obtained using the nearest distance to the point cloud rather than the predicted signed distance. The result of ``0'' indicates that weighting queries nearer to the surface more is important for the level set alignment, since all level sets are aligned to the zero level set. The result of ``Euclidean'' indicates that using inferred SDF achieves comparable results (the results of ``10'') with using its nearest distance to the point cloud, but finding the nearest point for each query may increase the computational burden in large scale point clouds. We also compare the decay parameters $\delta$ to obtain $\beta_{\bm{q}}$ in Eq.~\ref{eq:totallossweight}, and $\delta=10$ performs the best.

\begin{table}
\centering
\resizebox{\linewidth}{!}{
    \begin{tabular}{c|c|c|c|c|c}
     \hline
    Adaptive Weight $\beta_{\bm{q}}$ &0&1&10&100&Euclidean\\
    \hline
    CD$\times 100$&0.455&0.439&0.394&0.489&\textbf{0.391}\\
    NC&0.955&0.957&\textbf{0.958}&0.956&\textbf{0.958}\\
   \hline
   \end{tabular}}
   \vspace{-0.1in}
   \caption{Effect of adaptive weight $\beta_{\bm{q}}$.}
   \label{table:AdaptiveWeight}
\end{table}

\noindent\textbf{Consistency with Surface Points. }We further justify how we compute gradient consistency. We report results with maximizing consistency between gradients at queries and gradients at their nearest points on the surface, rather than their projections on the zero level set. Compared to the projections on the zero level set which is optimized in different iterations, the nearest point on surface is fixed. In Tab.~\ref{table:Loss Function}, the results of ``Fixed'' degenerate from the results of ``Cosine''. The reason is that, during the early stage of optimization, the surface may not be the zero level set of the learned SDF, which brings lots of ambiguity and conflict if we use the nearest point as a reference. Hence, using projections on the zero level set in current iteration produces better results.

\begin{table}
\centering
\resizebox{\linewidth}{!}{
    \begin{tabular}{c|c|c|c|c|c|c}
     \hline
    Loss&Fixed&MSE-Nor&MSE&Cosine&DiGS&DiGS+Cosine\\
    \hline
    CD$\times 100$&1.384&0.486&0.494&\textbf{0.394}&0.601&0.412\\
    NC&0.941&0.953&0.951&\textbf{0.958}&0.938&0.951\\
   \hline
   \end{tabular}}
   \vspace{-0.1in}
   \caption{Ablation studies on the loss function.}
   \label{table:Loss Function}
\end{table}

\noindent\textbf{Cosine Distance. }We show the advantages of cosine distance in Eq.~\ref{eq:consistency}. We replace cosine distance using a mean squared error with normalized gradients or with gradients without normalization. The results of ``MSE-Norm'' and ``MSE'' in Tab.~\ref{table:Loss Function} show that cosine distance performs better than MSE in SDF inference.

\begin{figure}[tb]
  \centering
  % the following command controls the width of the embedded PS file
  % (relative to the width of the current column)
  %\includegraphics[width=.95\linewidth, bb=39 696 126 756]{figures/definition3.eps}
   \includegraphics[width=\linewidth]{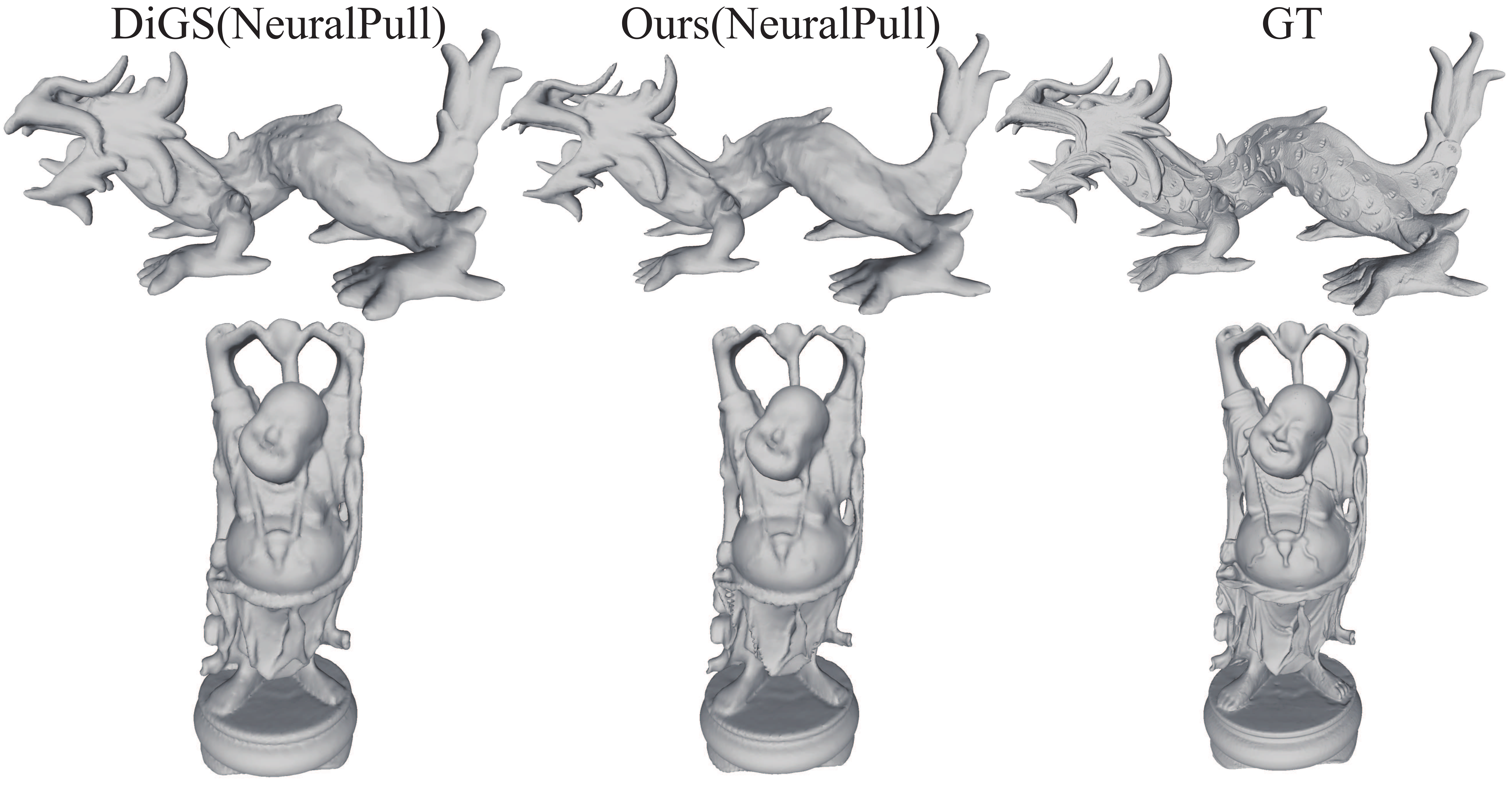}
  % replacing the above command with the one below will explicitly set
  % the bounding box of the PS figure to the rectangle (xl,yl),(xh,yh).
  % It will also prevent LaTeX from reading the PS file to determine
  % the bounding box (i.e., it will speed up the compilation process)
  % \includegraphics[width=.95\linewidth, bb=39 696 126 756]{sampleFig}
  %
  %
  \vspace{-0.2in}
\caption{\label{fig:DiGS}Visual comparison with the constraint on second order derivatives in DiGS.
}
\vspace{-0.1in}
\end{figure}

\noindent\textbf{Constraint on Second Order Derivatives. }We compare our loss with the constraints on second order derivatives in~\cite{ben2021digs} which aims to smooth the change of gradients. Although our loss also involves second order derivatives during gradient descent, we do not explicitly add constraints on the second order derivatives, which may result in unstable optimization. The comparison with the results of ``DiGS'' and the results of ``DiGS+Cosine'' indicate that our loss can reveal more accurate SDFs than the constraint on second order derivatives. The visual comparison with~\cite{ben2021digs} in Fig.~\ref{fig:DiGS} shows that the constraint on second order derivatives can not achieve more compact and sharper surfaces as ours.

%\noindent\textbf{Eikonal Term. }We show that our loss does not rely on the Eikonal Term which constrain gradients to be 1 everywhere in the field. This term ensures the predicted distances to be signed distances in methods like IGR~\cite{DBLP:conf/icml/GroppYHAL20}, SIREN~\cite{sitzmann2019siren} and NeuS~\cite{neuslingjie}, while NeuralPull predicts signed distances without the Eikonal term. Our loss can take effect upon a signed distance field rather than with the Eikonal term. The result of `'Eikonal'' indicates that the Eikonal term degenerates the performance of NeuralPull, even with our loss.

%1. 总得权重
%2. 每个点的权重， 预测sdf，不要这个权重， 用到表面最近距离
%2.1 把这个adaptive的权重去掉
%3. cosine变成l2 以及不归一化模长直接l2最小
%4. 拉完配对 还是和表面点配对
%5. np 不加长度等于1 igr siren加一 图片加一
%6. 和二阶偏导的对比

\section{Conclusion}
We improve the learning of SDFs without signed distance supervision by pursuing better gradient consistency. Our analysis shows that consistent gradients in the field are the key factor affecting the accuracy of inferred SDFs. To evaluate the gradient consistency, we introduce a level set alignment loss. By minimizing our loss, we successfully align all level sets onto the zero level set, which propagates the zero level set to eliminate 3D ambiguity through better gradient consistency. Our loss can be applied upon different methods a general term in loss function to improve the gradient consistency in the SDFs inferred from 3D point clouds or multi-view images. The visual and numerical comparisons with the state-of-the-art methods justify our effectiveness and show our superiority over the latest methods in SDF inference.

%%%%%%%%% REFERENCES
{\small
\bibliographystyle{ieee_fullname}
\bibliography{papers}
}

\end{document}